\documentclass[letterpaper, 10 pt, conference]{ieeeconf}  

\IEEEoverridecommandlockouts                              

\usepackage{graphics} 
\usepackage{epsfig} 
\usepackage{mathptmx} 
\usepackage{times} 
\usepackage{amsmath} 
\usepackage{amssymb}  
\usepackage[font=footnotesize,labelfont=bf]{caption}
\usepackage{subcaption}
\title{\LARGE \bf
Human-Robot Interaction in Retinal Surgery: A Comparative Study of Serial and Parallel Cooperative Robots
}

\author{Botao Zhao$^{1}$, 
Mojtaba Esfandiari$^{1}$,
David E. Usevitch$^{1}$, {\it Student Member, IEEE}, 
Peter Gehlbach$^{2}$, {\it Member, IEEE},  
\\Iulian Iordachita$^{1}$, {\it Senior Member, IEEE}
\thanks{*This work was supported by U.S. National Institutes of Health under the grants number 2R01EB023943-04A1 and 1R01 EB025883-01A1, and partially by JHU internal funds.}
\thanks{$^{1}$ Botao Zhao, Mojtaba Esfandiari, David E. Usevitch, and Iulian Iordachita are with the Department of Mechanical Engineering and Laboratory for Computational Sensing and Robotics, Johns Hopkins University,
Baltimore, MD, 21218, USA. 
        ({\tt\small bzhao17,mesfand2,usevitch, iordachita@jhu.edu})}%
\thanks{$^{2}$ Peter Gehlbach is with the Wilmer Eye Institute, Johns Hopkins Hospital, Baltimore, MD, 21287, USA. ({\tt\small pgelbach@jhmi.edu})
}
}

\begin{document}

\maketitle
\thispagestyle{empty}
\pagestyle{empty}

\begin{abstract}

Cooperative robots for intraocular surgery allow surgeons to perform vitreoretinal surgery with high precision and stability. Several robot structural designs have shown capabilities to perform these surgeries.
This research investigates the comparative performance of a serial and parallel cooperative-controlled robot in completing a retinal vessel-following task, with a focus on human-robot interaction performance and user experience.
Our results indicate that despite differences in robot structure and interaction forces and torques, the two robots exhibited similar levels of performance in terms of general robot-to-patient interaction and average operating time. These findings have implications for the development and implementation of surgical robotics, suggesting that both serial and parallel cooperative-controlled robots can be effective for vitreoretinal surgery tasks.

\end{abstract}


\section{INTRODUCTION}
Vitreoretinal surgery (VRS) is among the most delicate and demanding eye surgery procedures in medicine. VRS deals with disorders of the vitreous and retina. The vitreous is a clear gel-like substance that fills the space between the lens and retina providing nutrients and structure to the eye. The retina is a thin layer of tissue lining the back of the eye which houses photoreceptor cells to receive photons and send electrical and chemical signals to the brain for visual perception \cite{Nguyen2023}.
VRS is used to treat a wide variety of conditions including retinal detachments, macular holes, epiretinal membranes, diabetic retinopathy, and age-related macular degeneration.
Given the challenging nature of these procedures, various assistive eye surgical robots have been developed in recent years. These robots can be broadly categorized into three major classes: handheld robots, teleoperated robots, and cooperative-controlled robots \cite{vander2020robotic}.

Handheld VRS robots are miniaturized robotic instruments directly held by eye surgeons without external support to the robot. In the typical case, an active tool is attached to the robot as its end-effector. As general tool manipulation is controlled directly by the surgeon, these robots are very familiar and intuitive and require less effort to master. The Micron system of Carnegie Mellon University \cite{yang2014manipulator} is among the best known examples of a handheld robot designed for VRS.
Teleoperated VRS robots are pairs of controller-responder interfaces, where the responder is remotely manipulated by a surgeon using a joystick or other user interface to move the controller mechanism. Benefits of this approach include motion scaling control and the elimination of hand tremor during surgery, of which both features are particularly beneficial to eye surgery tasks. The SMOS \cite{guerrouad1989smos}, Preceyes \cite{meenink2013robot}, and RAMS \cite{nasseri2013introduction} robots are all notable examples of VRS teleoperated robots. 
Cooperative-control (CC) robots (“co-bots”) are designed to allow control by both the surgeon and robot. The surgeon holds a surgical instrument supported by the robot, connected via a force-torque sensor. As the surgeon moves the tool, the robot moves in real time according to assigned corresponding compliant motions mapped to forces and torques detected. These motions are dictated by specified robot kinematics and selected control algorithms. Co-bots can reduce or eliminate hand tremor while still maintaining the direct tool manipulation experience familiar to surgeons. Well-known examples of cooperative-control robots for VRS include the robotic systems of King’s College London and Moorfields Eye Hospital \cite{mablekos2018requirements}, KU Leuven \cite{gijbels2013design} and the Steady-Hand Eye Robot (SHER) developed by Johns Hopkins University \cite{he2012toward}.

\begin{figure}[b!]
  \centering
    \includegraphics[width=86mm]{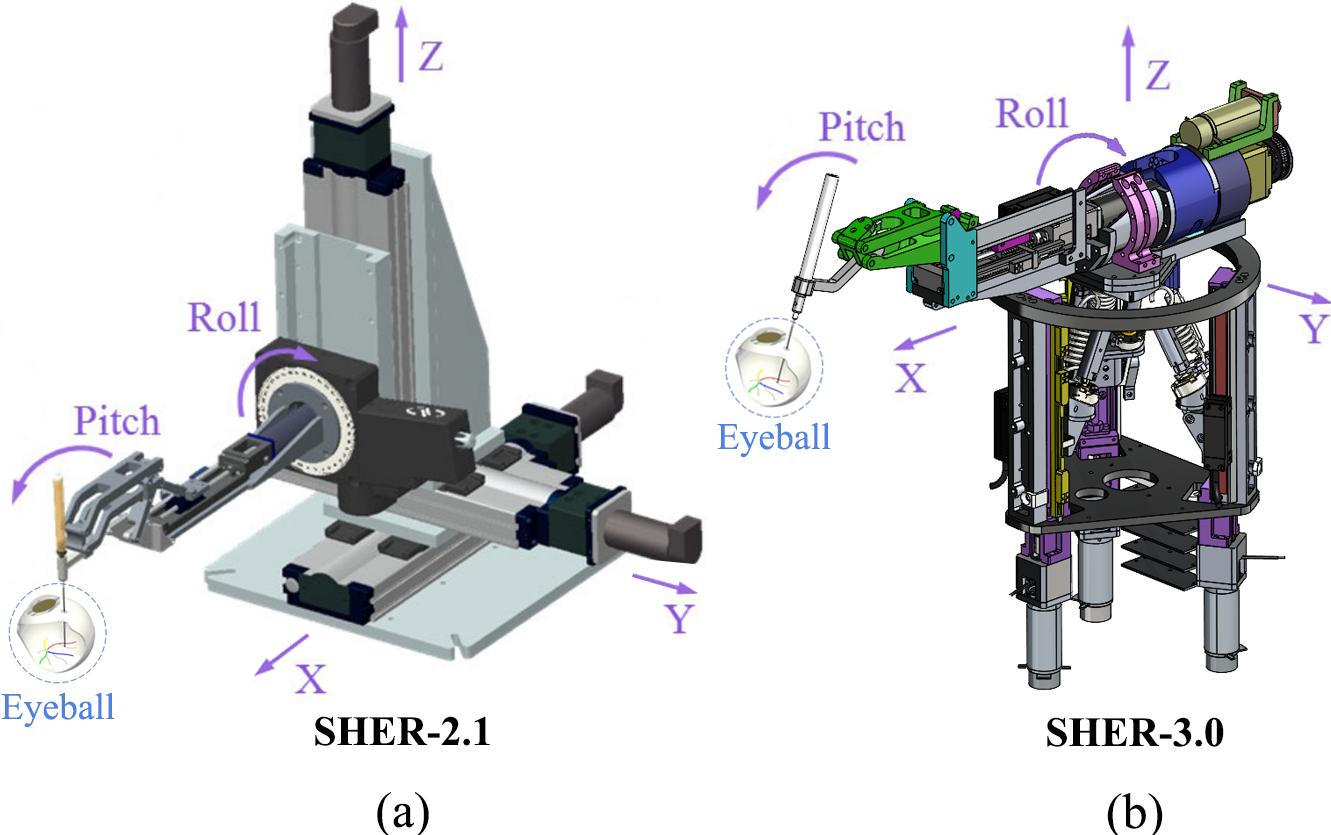}
      \caption{CAD models of (a) the cartesian serial robot: SHER-2.1, and (b) the delta parallel robot: SHER-3.0.}
      \label{fig: ER_CAD}
\end{figure}


VRS robotic systems have implemented various effective robotic designs to achieve precise motions for surgery tasks including the use of serial and parallel designs. Previous research has been devoted to the design, analysis, and comparison of robot kinematics, dynamics, and control systems for both serial and parallel robotic systems \cite{pandilov2014comparison},\cite{nzue2013comparison} though comparison of these two types is not a trivial matter. In general, serial robots demonstrate excellent repeatability and can be programmed with relative ease to perform a broad range of surgical procedures, but their larger size compared to parallel robots may pose challenges in certain surgical settings. Parallel robots are generally capable of faster dynamic motions, improved rigidity, and their ability to be restricted to a compact size makes them well-suited for use in tight surgical spaces. Parallel robot design and implementation, however, involves more complex mathematics, programming, and operation compared to serial robots.

Despite the extensive research in robot design and control outlying benefits and drawbacks between these two robot types, there is a lack of comparative studies in using them for medical surgery tasks. As comparing these robot types is quite complex with many factors coming into play (e.g., intricacies of kinematics, dynamics, Jacobian computations, and a slew of others), simplifying comparisons to focus primarily on human-robot interaction could prove valuable, as this component is the most essential when considering CC robots. Improved understanding of this interaction may show benefits for using a certain robot design for cooperative surgical tasks over the other. 
Hence a comparative study based on human-robot interaction could potentially reveal significance in performances between styles influencing further design trends for CC VRS robots of the future.

\begin{figure}[t!]
  \centering
    \includegraphics[width=80mm]{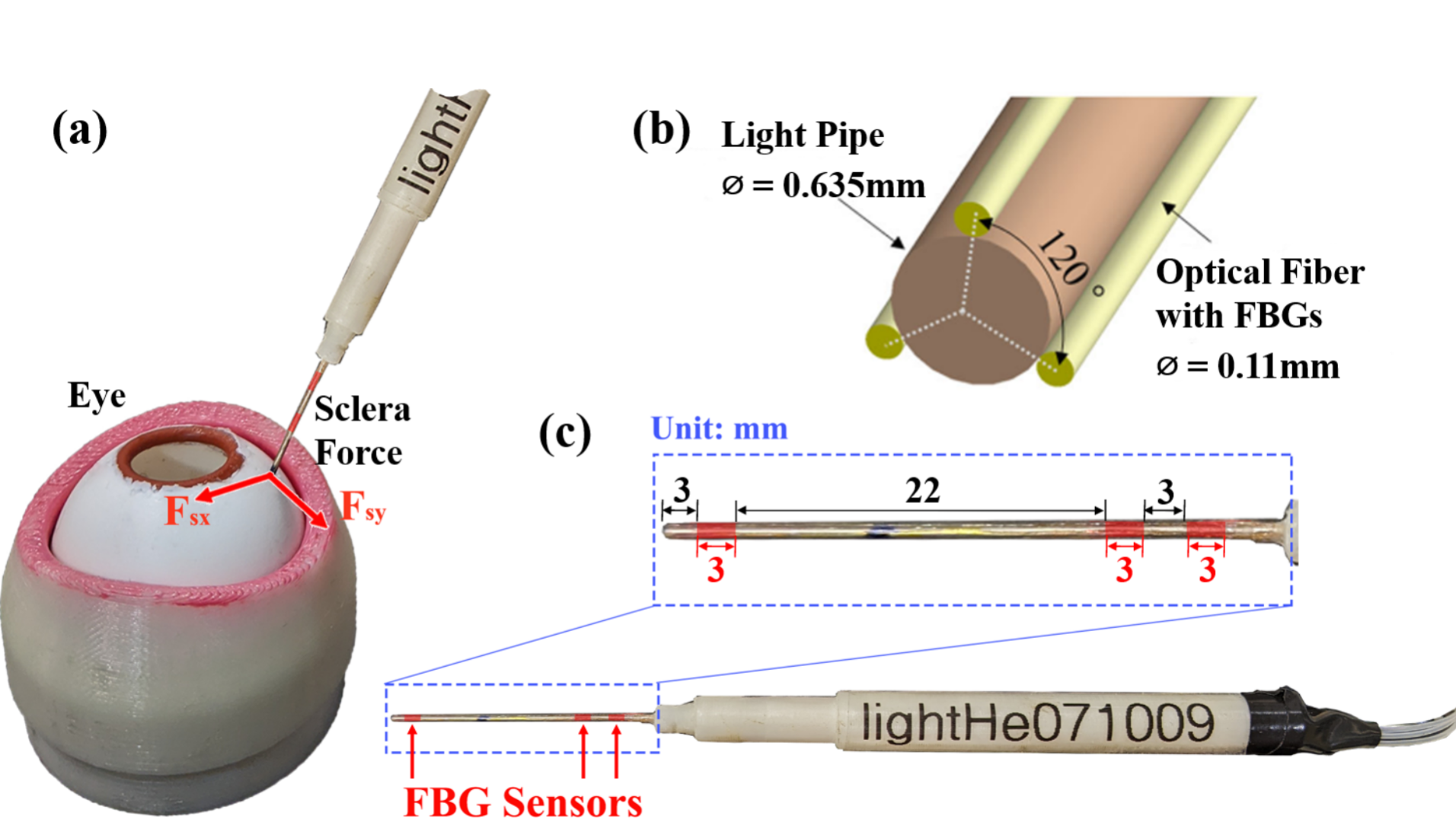}
      \caption{Design and dimensions of an FBG force sensing tool. (a) The tool shaft interacting with an eyeball phantom. (b) Section view and geometry of the tool shaft \cite{he2020automatic}. (c) The dimensions of the 3\,mm-long FBG sensors.}
      \label{fig: FBG_sensor}
\end{figure}

This work provides a comparative study between two kinematically different SHER robots (SHER-2.1 and SHER-3.0), with serial and parallel robot designs, to better understand their differences for VRS use. As the serial-based SHER-2.1 has previously been verified as a reliable and effective robot for VRS \cite{he2012toward},\cite{ebrahimi2018real}, it is used as the base reference for comparison with the SHER-3.0, the parallel robot, in this study. For human-to-robot interaction comparison, the force and torque of the user's hand-to-tool interaction and the tool-to-eye phantom interaction force of each robot are measured and recorded. Learning curves and the cumulative summation (CUSUM) technique \cite{nagakawa2018learning} are used to assess user proficiency and compared between manipulations of each robot.

This work contributes insight into the comparison of serial and parallel robots for human-to-robot interaction using CC VRS robots. We present a unique vessel following task which can be used to compare CC robot performance, and analysis methods for the comparison of human-to-robot and tool-to-tissue forces and torques from such robots.

\section{THE STEADY-HAND EYE ROBOT}

The Steady-Hand Eye Robot was designed for surgeons to synergically share tool control with the robot. Both hold the surgical instrument which is attached to the robot via a force-torque sensor. The tool handle transmits measured forces and torques applied by the user at the handle frame as shown in Fig.\ref{fig: Workflow}. An admittance control algorithm converts applied forces and torques to desired translational and rotational velocities. An optical fiber-based force sensor is embedded in the tool shaft to measure tool-to-eye interaction force.

\subsection{SHER-2.1}

SHER-2.1 is a 5 degree of freedom (DoF) Cartesian serial robot as shown in Fig.\ref{fig: ER_CAD}(a). The three prismatic actuators provide translation in the x, y, and z directions. The extension arm mounted on the vertical sliding rail has an additional 2-DoF of rotation about the x and y axes. The robotic extension arm includes a compact tool holder with a quick release mechanism for inserted surgical instruments, a stiff structural design, and generates a sufficient work space for tool rotation \cite{he2012toward}.

\subsection{SHER-3.0}

SHER-3.0 is a 5-DoF delta parallel robot as shown in Fig.\ref{fig: ER_CAD}(b). The delta system uses three vertical arranged prismatic actuators to move the delta arms. This provides translation along the x, y, and z directions while keeping the delta platform horizontal. The robotic extension arm mounted on the delta platform has an additional 2-DoF of rotational motion about the x and y axes like SHER-2.1. SHER-3.0, however, has a simplified four-bar linkage mechanism. This mechanism offered more hand space to surgeons due to a longer extension arm comparied to SHER-2.1, thus improving comfort and workspace limitations during use. The four-bar linkage mechanism is actuated by an offset slider-crank actuated by a rotary actuator \cite{roth2021towards}. As SHER-2.1 has been used successfully in previous studies, SHER-3.0 gains were tuned before any experimentation to demonstrate a similar response and feel to that of SHER-2.1 when moving a tool.

\begin{figure*}[t!]
  \centering
    \includegraphics[width=160mm]{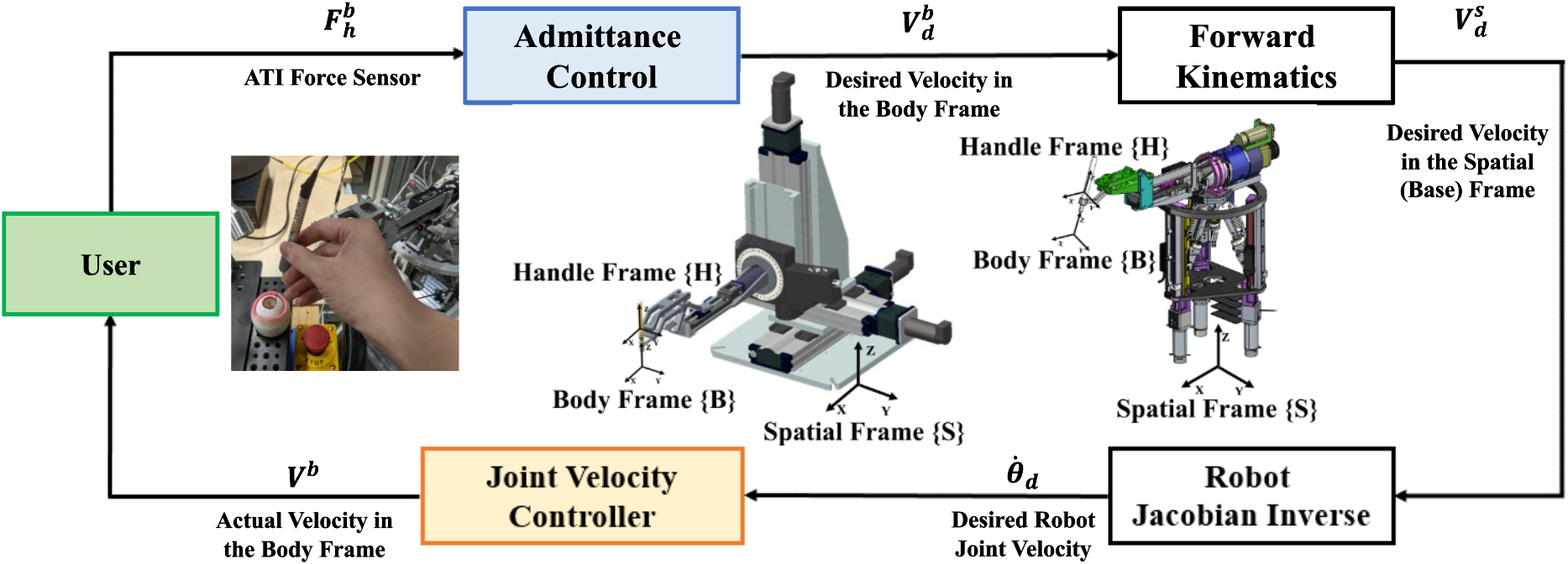}
      \caption{Block diagram showing the admittance control mode used in this experiments to synergically control the SHER. The transformation between the body frame \{B\} to the handle frame \{H\} is the same for both robots.}
      \label{fig: Workflow}
\end{figure*}
\begin{figure}[!b]
  \centering
    \includegraphics[width=75mm]{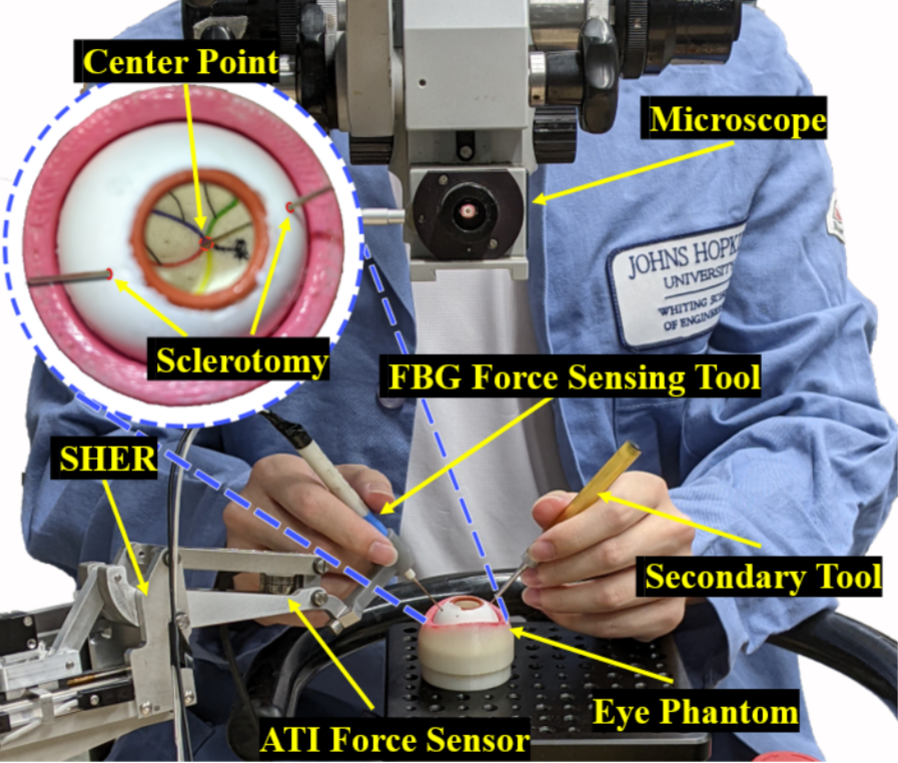}
      \caption{Experimental setup including the SHER, the ATI Nano17 force/torque sensor, the FBG-equipped force sensing tool, the secondary tool, the eye phantom, and the Zeiss surgical microscope.}
      \label{fig: manipulation}
\end{figure}

\subsection{FBG Force Sensing Tool}

A Fiber Bragg Grating (FBG) force sensing tool mounted on the surgical instrument is used for measuring amount of force applied to the tool directly from its interaction with eye tissue. \cite{he2020automatic}.
The tool shaft is made of a microsurgical light pipe with three FBG fibers (Technica S.A., Beijing, China) arranged at 120\textdegree{} intervals around the light pipe as shown in Fig. \ref{fig: FBG_sensor}b. Each fiber contains three separate 3\,mm-long FBG sensors as shown in Fig. \ref{fig: FBG_sensor}c.

\section{ADMITTANCE CONTROL SCHEME}



Both SHER robots employ an admittance control algorithm that allows surgeons to cooperatively manipulate the tool attached to the robot force-torque sensor, and guide the SHER towards a desired configuration. The 6-DoF sensor outputs a wrench vector $F_h^b =[f_b,\tau_b]^T\in \mathbb{R}^6$, which includes the hand-to-tool interaction force $f_b\in \mathbb{R}^3$ and torque $\tau_b \in \mathbb{R}^3$ exerted to the robot handle, represented in the robot body frame as shown in Fig. \ref{fig: Workflow}. This wrench $F_h^b$ is then used in the admittance control algorithm to generate a desired end-effector velocity vector in the robot body frame denoted as $V_d^b = [v_d , \omega_d]\in \mathbb{R}^6$ as follows
\begin{equation} 
V_d^b = \mathbb{K} F_h^b 
\label{eq: admittance_law}
\end{equation}
in which $\mathbb{K}\in \mathbb{R}^{6 \times 6}$ is a diagonal matrix of constant gains, and $v_d$ and $\omega_d$ are the desired end-effector linear and angular velocities respectively, represented in the robot body frame. Equation \ref{eq: admittance_law} shows that the desired end-effector velocity $V_d^b$ is proportional to the hand-tool interaction force $F_h^b$ 
and thus the level of proportionality for each DoF can be separately adjusted by the magnitude of the diagonal entries of matrix $\mathbb{K}$.   

Once the desired end-effector velocity $V_d^b$ is found in task space, the desired joint space angular velocity vector can be computed as follows
\begin{equation} 
\Dot{\Theta}_d = J_{sb}^\dagger(\Theta)V_d^b 
\label{eq: joint_velocities}
\end{equation}
where $\Theta \in \mathbb{R}^5$ denotes the joint variables, $\dot{\Theta}_d \in \mathbb{R}^5$ is the desired joint space angular velocity, $J_{sb}\in \mathbb{R}^{6 \times 5}$ is the robot Jacobian which maps joint velocities to end-effector velocities between the robot spatial frame (shown in Fig. \ref{fig: Workflow}) and the robot body frame, and $J^{\dagger}_{sb}\in \mathbb{R}^{5 \times 6}$ is the Jacobian pseudoinverse which maps the end-effector velocity to joint velocities and is computed below as in \cite{chiacchio1991closed}.
\begin{equation}
J^{\dagger}_{sb} = J^T_{sb}(J_{sb}J^T_{sb})^{-1}
\label{eq: Jscobian_pseudo_inverse}    
\end{equation}
At each defined time step the low-level motor controller generates desired joint velocities $\dot{\Theta}_d$ to command the robot motors for corresponding end-effector velocities $V_d^b$ comeing from the computed Jacobians.


\begin{figure*}[!t]
  \centering
    \begin{subfigure}[b]{0.45\textwidth}
        \centering
        \includegraphics[width=\textwidth]{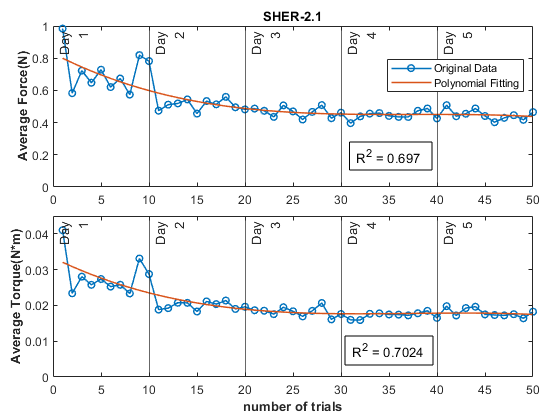}
        \caption{}
        \label{fig: Learn_Curve_2.1}
    \end{subfigure}
    \begin{subfigure}[b]{0.45\textwidth}
        \centering
        \includegraphics[width=\textwidth]{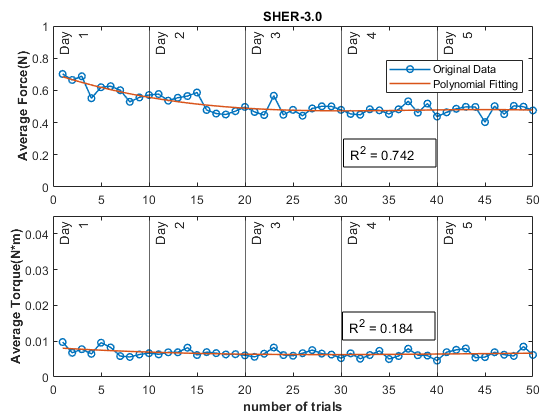}
        \caption{}
        \label{fig: Learn_Curve_3.0}
    \end{subfigure}
    \centering
        \begin{subfigure}[b]{0.45\textwidth}   
            \centering 
            \includegraphics[width=\textwidth]{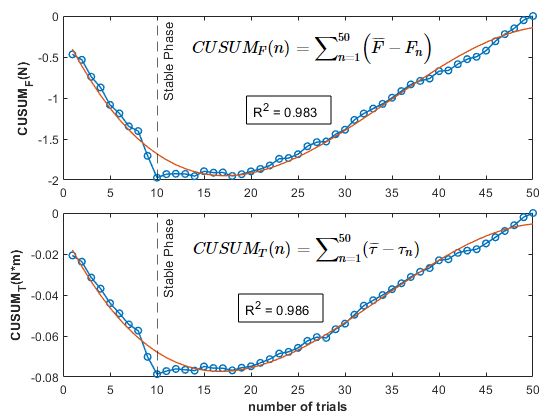}
            \caption{}
            \label{fig: CUSUM_2.1}
        \end{subfigure}
        \begin{subfigure}[b]{0.45\textwidth}   
            \centering 
            \includegraphics[width=\textwidth]{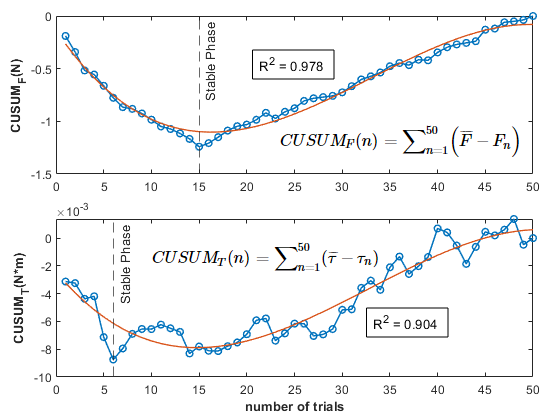}
            \caption{}
            \label{fig: CUSUM_3.0}
        \end{subfigure}    
    \caption{The original data points with third-degree polynomial fitting. (a) The learning curves of the user's average hand-to-tool interaction force and torque using the SHER-2.1. (b) The learning curves of the user's average hand-to-tool interaction force and torque using the SHER-3.0. (c) The CUSUM curves of handle force and torque using the SHER-2.1. (d) The CUSUM curves of handle force and torque using the SHER-3.0. For better visualization results, (c) and (d) are plotted in different scales.
    }
\end{figure*}

\section{EXPERIMENT}
An pilot-study was designed to quantify the cooperative-control performance between the user and each SHER by analyzing both user hand-to-tool interaction forces and torques, and tool-to-eye interaction forces for a vessel-following task. A right-handed, non-clinical user was used to collect data.

\subsection{Experimental Setup}

Both experimental setups are depicted in Fig.\ref{fig: manipulation}, which included either SHER, an ATI force-torque sensor (Nano17, ATI Industrial Automation, NC, USA), an FBG-equipped force-sensing tool with the interrogator (Technica Optical Components, China), a silicone eye phantom, a secondary stabilization tool, and a Zeiss surgical microscope. A desktop computer was connected to the SHERs, force-torque sensor, and FBG-equipped tool using a TCP-IP connection.

The silicone eye phantom shown in Fig.\ref{fig: manipulation} with four colored vessels painted onto the retina was placed under the surgical microscope for the user to view during tool manipulation. Eye phantoms are anatomically-similar eye models commonly used during benchtop experiments that imitate some or many properties of the human eye \cite{gupta2014human}.

\begin{table*}[t!]
\centering
    \caption{The comparison results of the ten random trials. A p-value less than 0.05 is considered to be statistically significant.}
    \includegraphics[width=160 mm]{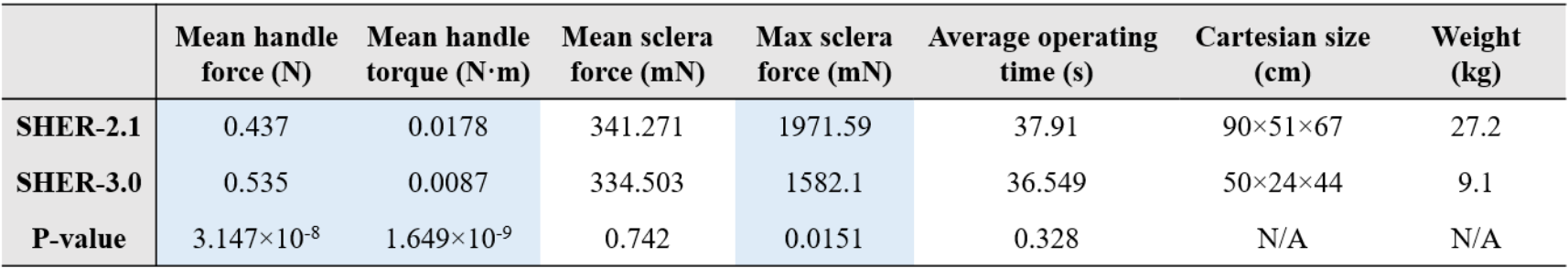}
      \label{table1}
\end{table*}

\begin{figure}[t!]
  \centering
    \includegraphics[width=84mm]{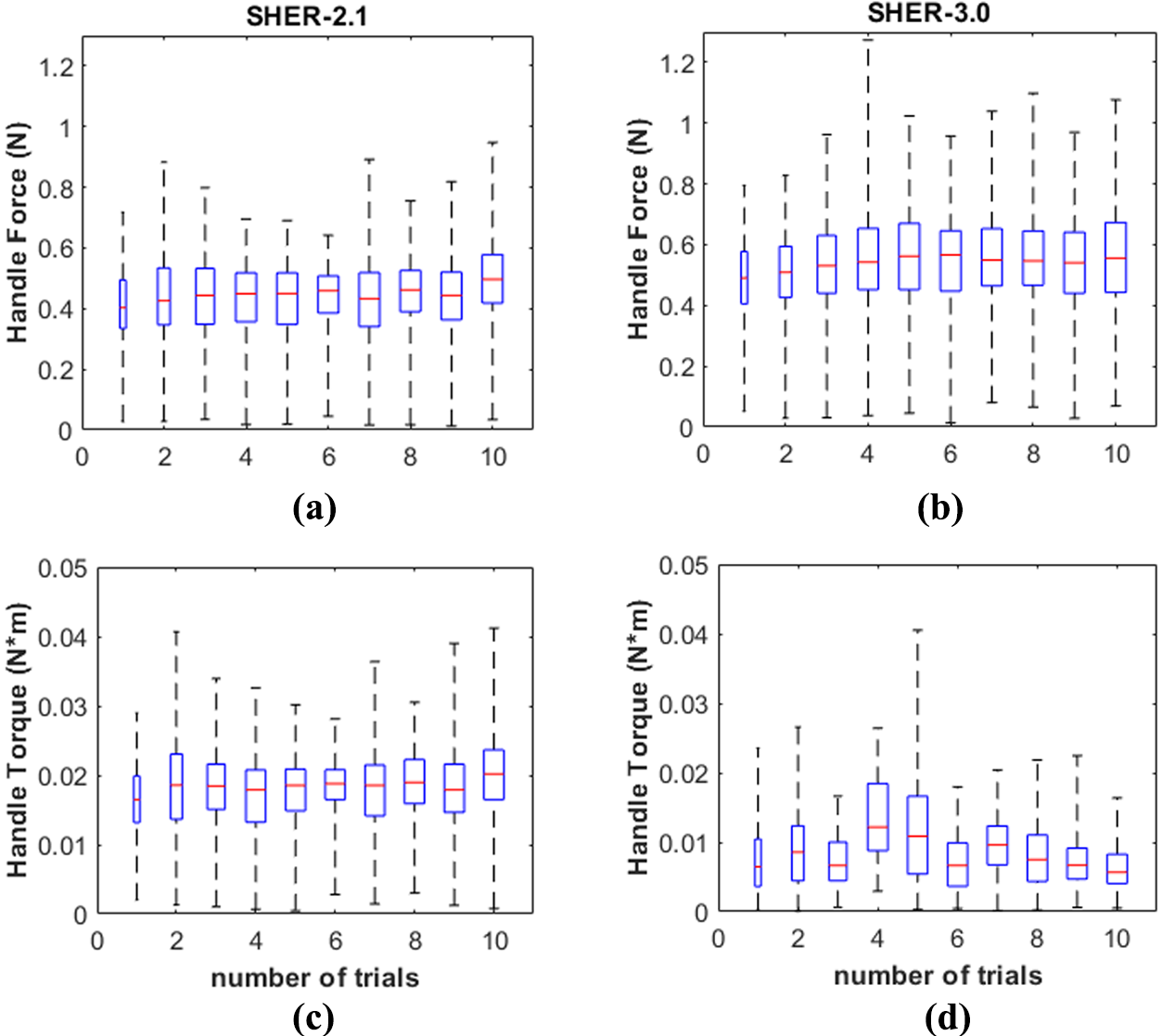}
      \caption{Data distribution of (a) hand-to-tool interaction force for the SHER-2.1 (b) hand-to-tool interaction force for the SHER-3.0 (c) hand-to-tool interaction torque for the SHER-2.1 (d) hand-to-tool interaction torque for the SHER-3.0}
      \label{fig: box_handle}
\end{figure}
\begin{figure}[t!]
  \centering
    \includegraphics[width=86mm]{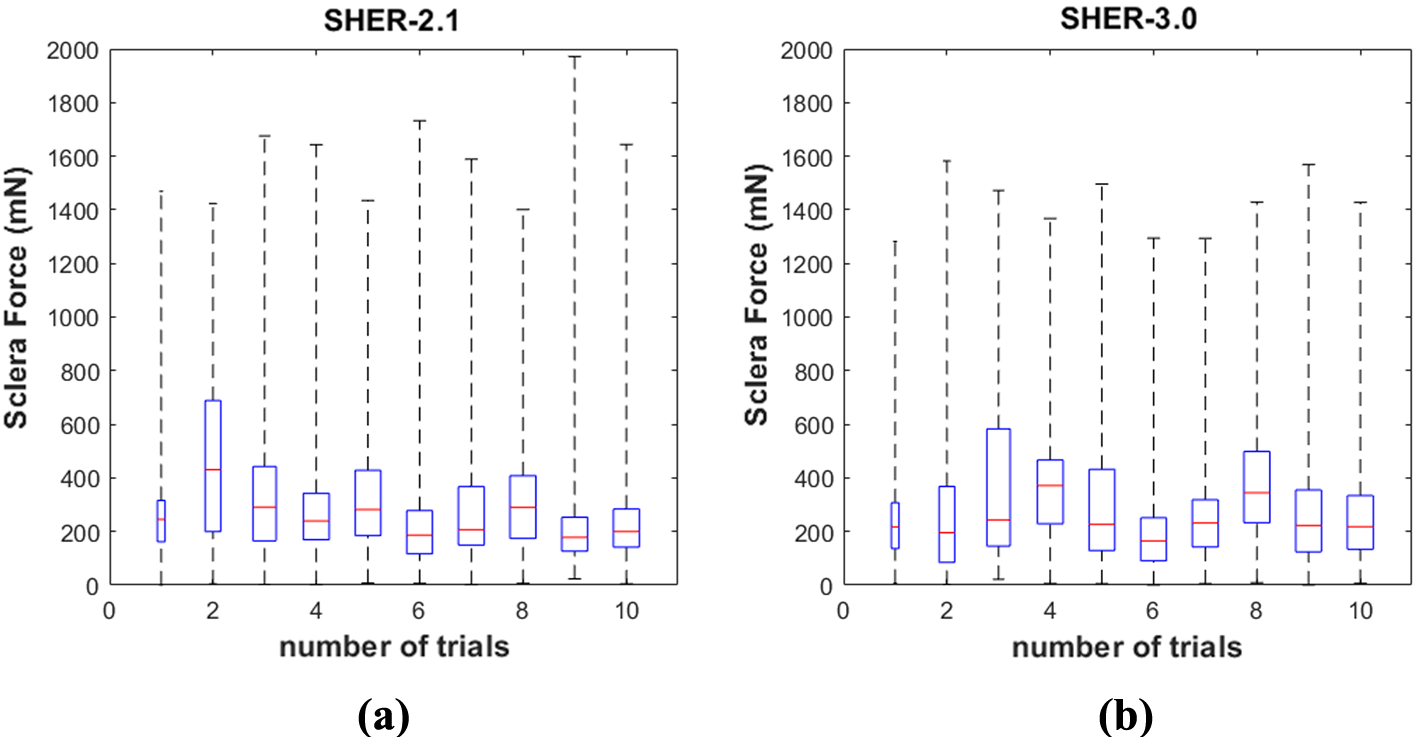}
      \caption{Data distribution of (a) tool-to-eye interaction force for the SHER-2.1 (b) tool-to-eye interaction force for the SHER-3.0}
      \label{fig: box_sclera}
\end{figure}

\subsection{Experimental Procedure}
To simulate VRS tasks, the user held the tool shaft attached to the SHER and inserted the tool into the eye phantom through a sclerotomy made on the phantom sclera tissue. The user then moved the tool tip along with the attached CC robots. Both robots used the previously described admittance control during operation. 

The user was tasked to follow four colored lines mimicking eye vessels with the tip of the FBG force-sensing tool. This vessel-following task is a simplified action, inspired from motions used in retinal membrane peeling where clinicians inspect the retina, move the eye, and navigate tools to specific locations to grab tissue.
Before starting each trial, the tool tip was placed "home" on the center of the circle (shown in Fig.\ref{fig: manipulation}) at the intersection point of the four colored vessels. The user then followed a vessel one at a time,  attempting to keep the central axis of the tool inline with the center axis of the followed vessel at all times. 
A secondary tool was used by the user to better control manipulation of the phantom eye for stabilization or rotation during vessel following. 

After the FBG force sensing tool was inserted into the eyeball through the sclerotomy, the user moved the tool tip along each vessel until reaching the end of the vessel, then backtracked to the starting point along the same path. This "moving back and forth" procedure was repeated until the user finished the trajectory on all four vessels. Completion of all four vessel vessel following tasks was recorded as one successful trial. To eliminate potential bias, the color sequence order was randomized for each trial.

To ensure the user had similar operating proficiency on both robots, 50 practice trials were conducted by the user on each robot prior to official experimentation. To quantify user proficiency, learning curves were constructed by plotting the average value of handle forces and torques of each trial. 
According to learning curve theory, as the the number of practice trials increases, eventual convergence of the performance curve should occur. We employed CUSUM analysis to assess the number of trials required for a participant to achieve a stable phase. CUSUM is a statistical method used to detect changes in the mean value of a process over time, and is often employed to monitor individual or group learning progress \cite{nagakawa2018learning}. CUSUM is useful in determining whether a participant has attained a stable level of performance, with a turning point on the CUSUM curve indicating a significant improvement in performance. In this study, the CUSUM curve was used as the performance curve, and defined as $CUSUM(n) = \sum_{n=1}^{50} (\bar{X}-X_n)$, where $\bar{X}$ was the mean value of overall force or torque of the 50 trials, and $X_n$ was the mean value of force or torque of the $n^{th}$ trial. After the learning phase, the user was asked to perform 10 trials of vessel following using each robot. To further eliminate the effect of uncertainties, the user conducted 10 trials with SHER-2.1 in the morning and with SHER-3.0 in evening and performed the same task on the next day in reverse order.

\begin{figure}[t!]
  \centering
    \includegraphics[width=81mm]{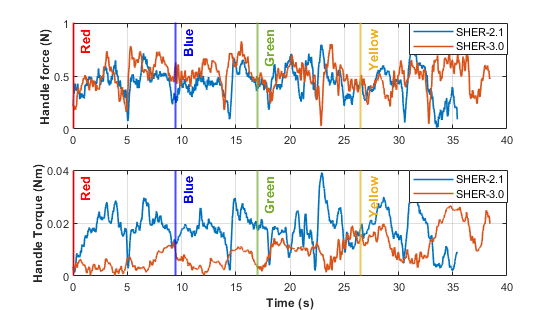}
      \caption{Handle force and torque magnitude variations for a single trial of each experiment set for the SHER-2.1 and the SHER-3.0.}
      \label{fig: RBGY_FT}
\end{figure}
\begin{figure}[t!]
  \centering
    \includegraphics[width=80mm]{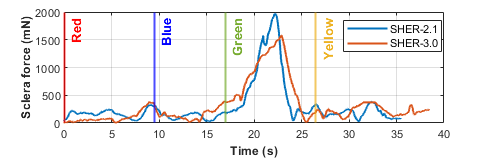}
      \caption{Scleral force magnitude variations for a single trial of the SHER-2.1 and the SHER-3.0. 
      }
      \label{fig: RBGY_sclera}
\end{figure}
\begin{figure}[t!]
  \centering
  \hspace*{1.5mm}
    \includegraphics[width=86mm]{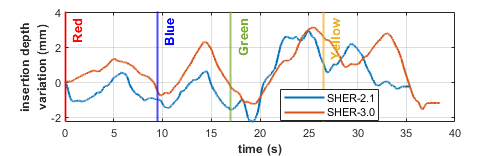}
      \caption{Insertion depth variations for a single trial of the SHER-2.1 and the SHER-3.0.}
      \label{fig: insertion_depth_variation}
\end{figure}
\begin{figure}[t!]
  \centering
    \includegraphics[width=80mm]{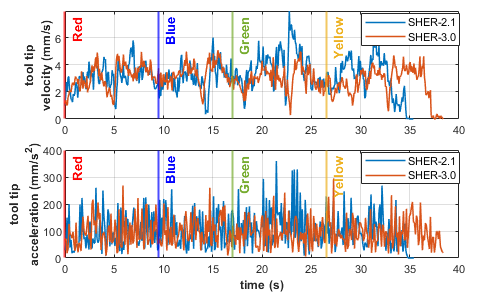}
      \caption{The magnitude of tool tip velocities and accelerations for a single trial of the SHER-2.1 and the SHER-3.0.}
      \label{fig: tool_tip_velocity_accleration}
\end{figure}
\begin{figure}[t!]
  \centering
    \includegraphics[width=80mm]{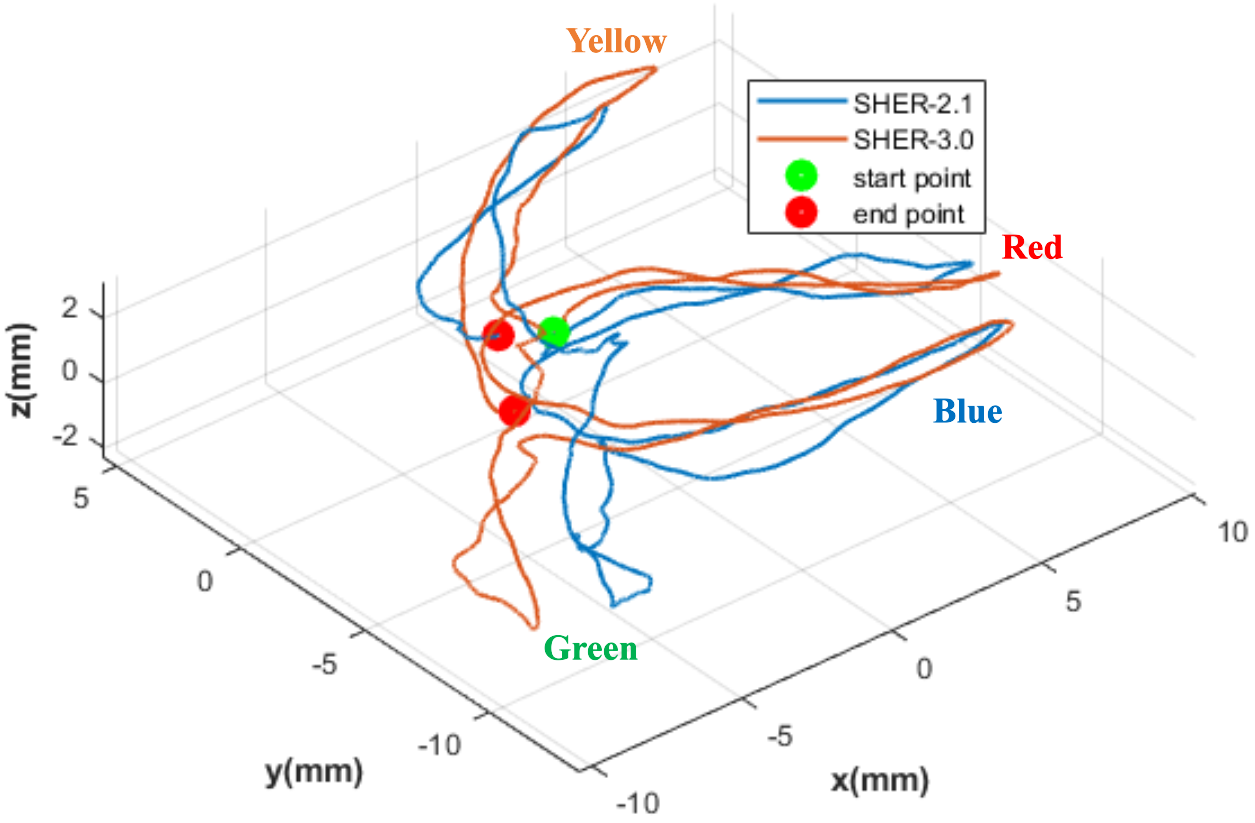}
      \caption{Tool tip trajectories for a single trial calculated from the forward kinematics of the SHER-2.1 and the SHER-3.0.}
      \label{fig: tool_tip_trajectory}
\end{figure}

\section{RESULTS AND DISCUSSION}

The average values of handle force and torque for each trial of the SHER-2.1 and the SHER-3.0 are plotted in Fig. \ref{fig: Learn_Curve_2.1} and Fig. \ref{fig: Learn_Curve_3.0} for visualizing the user's level of proficiency on each robot. Seen is a negative correlation between the number of trials and the hand-to-tool interaction force and torque for both robots. This shows tool forces decrease over time as the user learns to control the tool using the SHER. Force and torque values converged within the 50 trial learning stage demonstrating proper learning of each SHER by the user before the following experiments. The notable turning point is identified as the lowest point occuring in the CUSUM curve. In this experiment, the turning point marks the boundary between the initial and stable phases of the learning curve. 
This is seen in Figs. \ref{fig: CUSUM_2.1} and \ref{fig: CUSUM_3.0} as the turning points line up well with the start of the stable phases of the force and torque learning curves seen in \ref{fig: Learn_Curve_2.1} and \ref{fig: Learn_Curve_3.0} respectively for each SHER.
Fig. \ref{fig: CUSUM_2.1} shows both the hand-to-tool interaction forces and torques reached the plateau of the learning curve after 10 trials using SHER-2.1. Similarly for the SHER-3.0, Fig.\ref{fig: CUSUM_3.0} shows the user's hand-to-tool interaction forces and torques reached a stable level after 15 trials and 6 trials respectively. Therefore, based on the results from the CUSUM graphs and fitted polynomial curves, we see the user reach a stable level of proficiency on both robots after 15 trials without significant change in later trials.

Fig. \ref{fig: box_handle} shows box plots representing distributions of handle torque and force data for all time intervals of the vessel following task during the experiment of 10 trials comparing both robots.
Fig. \ref{fig: box_handle}a and Fig. \ref{fig: box_handle}b display handle forces, showing the distribution of the median values of the handle forces for the two robots fluctuated around 0.4\,N and 0.5\,N respectively. Fig. \ref{fig: box_handle}c and Fig. \ref{fig: box_handle}d display handle torques, showing the median values for SHER-2.1 to be about 0.015\,Nm, which is about double that for SHER-3.0. Thus the user is required to apply greater torques to change the orientation of the SHER-2.1 tool  as compared to when using SHER-3.0. The different extension arm length most likely impacted these results.

Fig. \ref{fig: box_sclera} shows box plots representing distributions of scleral forces for all time intervals of the vessel following task during the experiment of 10 trials comparing the SHER-2.1 and the SHER-3.0. Most values are concentrated between 100\,mN and 300\,mN with no observable shift in the distribution of sclera force among the 10 trials. This indicated that the scleral insertion force for both robots remained consistent across all trials.

Table \ref{table1} summarizes results from the 10 trials of vessel following using the two robots. The table also includes information on the dimensions and weights of the robots. The overall Cartesian size of each robot was measured as X$\times$Y$\times$Z as shown in Fig. \ref{fig: ER_CAD}, with length, width, and height measured when the robots were in their homed configurations.
A two-sample t-test was performed to test the null hypothesis that the two data sets would have have no significant differences between their mean handle forces and torques, mean and max scleral forces, and average operating times \cite{montgomery2017design} using a \emph{P}-value of 0.05 to determine significance. The average user-robot interaction force values for SHER-2.1 were significantly lower than those using SHER-3.0 (0.437\,N and 0.535\,N respectively, \emph{P}=$3.147\times10^{-8}$). The user applied a significantly smaller average torque using SHER-3.0 than using SHER-2.1 (0.0178\,Nm versus 0.0087\,Nm, \emph{P}=$1.649\times10^{-9}$).
Importantly, the SHER-3.0 also had a significantly smaller maximum scleral interaction force (1582.1\,mN versus 1971.59, \emph{P}=$0.0151$) although the mean scleral force, average operating time, and procedure time p-values were not found to be significantly different. Although there was variability in individual trials and some significant differences experienced between the two robots, the overall performance using both robots was generally similar. Though the reduction of maximum scleral forces is ideal as demonstrated by SHER-3.0 it is difficult to conclude that SHER-3.0 outperformed SHER-2.0 (or vice versa) though significant differences were also found between mean handle forces and torques.

Fig. \ref{fig: RBGY_FT} shows the user hand-to-tool interaction force and torque during a single trial, using the two robots to perform the standard vessel following sequence: Red-Blue-Green-Yellow (RBGY). The handle torque of the SHER-2.1 can be seen is noticeably larger than the SHER-3.0 in this trial, though handle torque differences are not so apparent. 

Fig. \ref{fig: RBGY_sclera} shows the variation of the scleral forces that each robot tool applied to the eye phantom during a single trial corresponding to Fig. \ref{fig: RBGY_FT}. The scleral force of the SHER-2.1 is seen generally consistent with the SHER-3.0 during operative time but with larger maximum forces. It should be noted that the scleral forces always reached the maximum allowed peak when the user followed the green vessel. This was due to the position of the insertion point was too close to the green vessel. This forced the user to use a steep angle to follow this vessel, leading to applying a larger force to the eye phantom. During eye surgery, this situation can be avoided by changing the scleral insertion point.

Fig. \ref{fig: tool_tip_trajectory} visualizes the calculated tool tip trajectories (assuming the tool is a rigid body) using both robots for a single trial along RBGY vessel paths. 
Fig. \ref{fig: insertion_depth_variation} shows the insertion depth variations on the z-axis, where the zero value represents the initial insertion depth of the tool at the start point.
Fig. \ref{fig: tool_tip_velocity_accleration} shows the magnitude of the tool tip velocity and acceleration for the single trial RBGY of the two robots.
These characteristics suggest consistency in user interaction between the SHER-2.1 and the SHER-3.0.

\section{CONCLUSION AND FUTURE WORK}

This study presents a preliminary comparative study of standardized vessel following performance while using both a CC serial robot (SHER-2.1) and parallel robot (SHER-3.0). 
Three major differences between the two robots were identified by these methods and include the mean handle torque, the mean scleral force, and the maximum sclera force applied to the eye phantom
during the operation. Based on the experimental results, we can conclude that with the same amount of training effort spent using both robots, the two robots had similar performance during a standardized vessel-following task. 
It's worth noting that SHER-3.0 is smaller and lighter than the SHER-2.1, making it easier to deploy in operating rooms. Given these findings, it is promising to assert that the SHER-3.0 holds great potential as a next-generation eye surgical platform although more work and surgery-specific tuning may improve its results.

Future comparative studies could include real-time tool tip trajectory analysis with computer vision algorithms possibly applying virtual boundaries to improve following outcomes. Additionally, increasing the number of users in the study could help validate and generalize results. This would ideally evaluate outcomes between non-clinical personnel compared to experienced eye surgeons. 
A proper study comparing outcomes between surgeon with various experience, would also be interesting and valuable for future exploration of CC surgical robots. Testing a task for targeting various positions in the retina (i.e. touching an array of grid points using both robots) would also be valuable for evaluating the robots for tasks such as retinal membrane peeling.
\newline

\addtolength{\textheight}{-11cm}   

\bibliographystyle{IEEEtran}
\bibliography{Bibliography}

\end{document}